\documentclass[conference]{IEEEtran}
\IEEEoverridecommandlockouts

\usepackage{cite}
\usepackage{amsmath,amssymb,amsfonts}
\usepackage{algorithmic}
\usepackage{graphicx}
\usepackage{textcomp}
\usepackage[table]{xcolor}
\usepackage{booktabs}
\usepackage{url}
\usepackage{multirow}
\def\BibTeX{{\rm B\kern-.05em{\sc i\kern-.025em b}\kern-.08em
    T\kern-.1667em\lower.7ex\hbox{E}\kern-.125emX}}
\begin{document}

\title{TESU-LLM: Training Speech-LLMs Without Speech via Unified Encoder Alignment}
\author{
  \IEEEauthorblockN{Taesoo Kim\IEEEauthorrefmark{1}\IEEEauthorrefmark{2} \quad Jong Hwan Ko\IEEEauthorrefmark{2}} 
  \IEEEauthorblockA{
    \IEEEauthorrefmark{1}KT Corporation, Republic of Korea\\
    \IEEEauthorrefmark{2}Department of Electrical and Computer Engineering, Sungkyunkwan University, Republic of Korea\\
    \texttt{ji5u1031@g.skku.edu}, \texttt{jhko@skku.edu}
  }
}

\maketitle

\begin{abstract}
Recent advances in speech-enabled language models have shown promising results in building intelligent voice assistants.
However, most existing approaches rely on large-scale paired speech-text data and extensive computational resources, which pose challenges in terms of scalability and accessibility.
In this paper, we present \textbf{TESU-LLM}, a novel framework that enables training speech-capable language models using only text data.
Our key insight is to leverage a unified encoder that maps semantically equivalent text and speech inputs to a shared latent space. By aligning the encoder output with the embedding space of a LLM via a lightweight projection network, we enable the model to generalize from text-only supervision to speech-based inference.
Despite being trained exclusively on text, TESU-LLM achieves strong performance on various speech-related benchmarks, comparable to baseline methods trained with large-scale multimodal datasets and substantial computational resources.
These results highlight the effectiveness and efficiency of our approach, offering a scalable path toward building speech LLMs without speech data.
\end{abstract}


\section{Introduction}

Large Language Models (LLMs) have shown impressive capabilities in understanding language, reasoning, and following complex instructions.
Recently, extending these models to support spoken language inputs has garnered increasing interest, aiming to power voice-based AI assistants that are both capable and versatile. However, building speech-capable LLMs typically demands large-scale paired speech-text datasets and significant computational resources~\cite{defossez2024moshi, abouelenin2025phi, chu2024qwen2, li2025baichuan}.
This high cost of data collection and training hinders accessibility and broad deployment, particularly in low-resource settings.

To alleviate these challenges, prior studies have explored more efficient training strategies.
For instance, parameter-efficient tuning methods~\cite{fang2024llama, wang2024freeze, wang2023blsp} reduce the number of trainable parameters to lower compute costs.
Other approaches~\cite{wang2025inserter, held2024distilling} propose distillation and pretraining pipelines that avoid explicit speech instruction data but still rely on raw speech during training.
A different line of work~\cite{dao2025speechlessspeechinstructiontraining} avoids both paired data and TTS, generating semantic speech tokens from text and aligning them to a frozen speech encoder.
While these approaches improve efficiency, they often suffer from performance gaps when compared to models trained on fully multimodal corpora~\cite{chu2024qwen2, abouelenin2025phi, li2025baichuan}.

Separately, multimodal foundation models like SpeechT5~\cite{ao2021speecht5}, SLAM~\cite{bapna2021slam, bapna2022mslam}, and SeamlessM4T~\cite{barrault2023seamlessm4t} demonstrate the power of unified encoder architectures in aligning speech and text representations within a shared latent space.
Though primarily designed for translation or speech synthesis tasks, their success indicates that unified modality representation may provide a scalable pathway for speech-language modeling—potentially reducing the reliance on paired data and explicit alignment techniques.

{In this work, we propose \textbf{T}ext-\textbf{E}nhanced \textbf{S}peech \textbf{U}nderstanding LLM (TESU-LLM), a simple yet effective framework for training speech-capable LLMs using only text supervision.
The key idea is that semantically equivalent speech and text inputs should be mapped to similar latent representations when processed through a shared encoder.
To realize this, we adopt a Unified Text-Speech Encoder that processes both speech and text inputs into a common latent space.
A lightweight encoder projector then maps these embeddings into the input space of a pretrained LLM.
This setup allows the model to generalize to spoken queries at inference, even though training is performed solely on text-based corpora—including both instruction-tuning datasets and synthetic long-form samples generated from a base LLM.

Extensive experiments show that TESU-LLM delivers strong performance across a diverse set of speech benchmarks, rivaling models that require large-scale speech-text corpora and massive compute resources.
This demonstrates the potential of our modality alignment approach as a scalable, compute-efficient solution for speech-language modeling.

\vspace{0.5em}
Our main contributions are summarized as follows:
\begin{itemize}
    \item We propose a novel framework for training speech-capable LLMs using only text-based data, without requiring speech data or any paired speech-text supervision.
    \item We leverage a unified speech-text encoder to generate speech-aligned representations directly from text, enabling interleaved training without the need for time-aligned segmentation or TTS synthesis.
    \item The proposed method trains only a lightweight encoder projector, achieving competitive results on speech-related tasks while using minimal GPU resources.
\end{itemize}
\begin{figure}[t]
    \centering
    \includegraphics[width=0.6\linewidth]{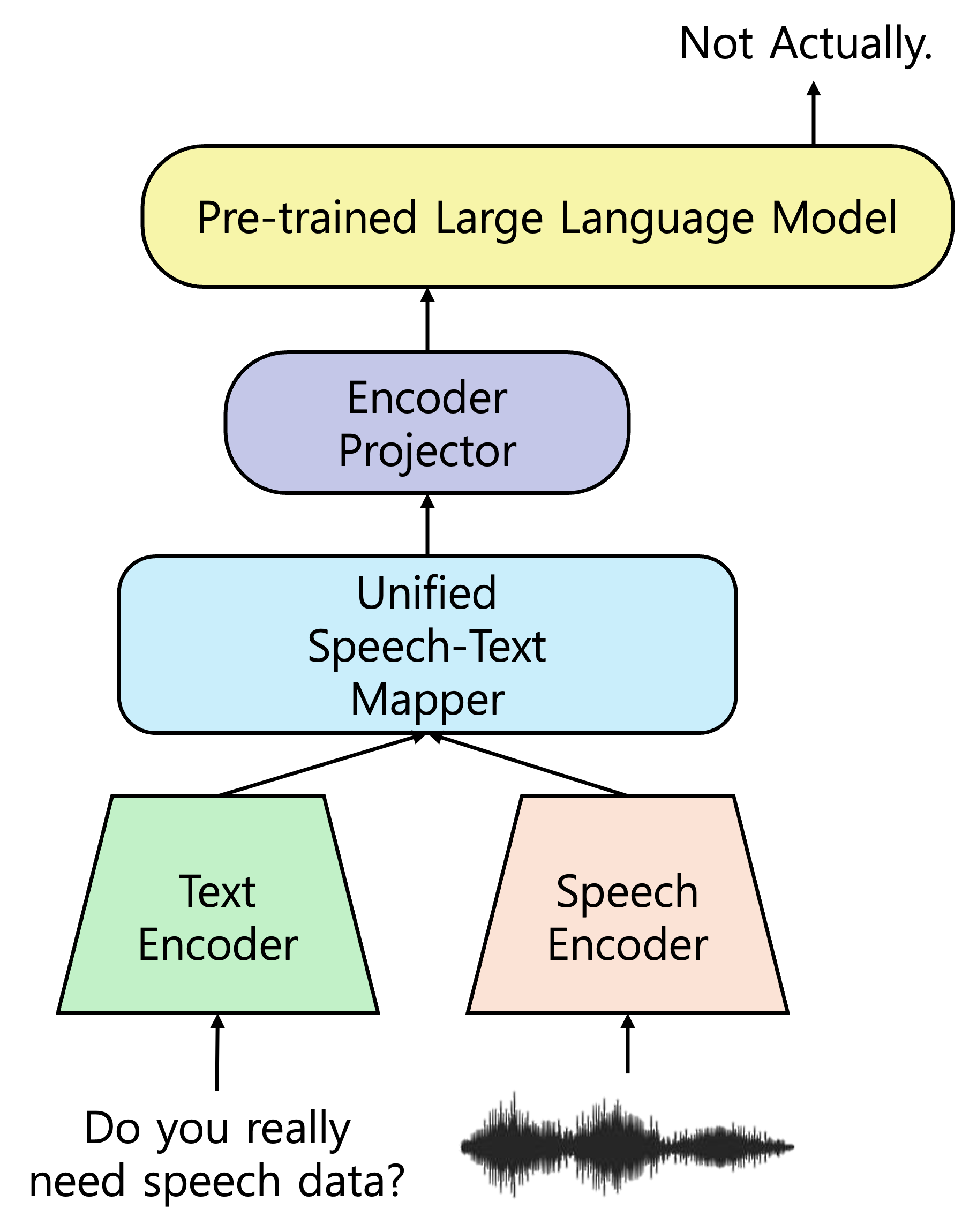}
    \caption{Overview of TESU-LLM architecture. The model consists of a Unified Text-Speech Encoder, an Encoder Projector, and a Pre-trained LLM. Text and speech inputs are encoded separately and mapped into a shared latent space, which is then projected to align with the LLM input. Only text is used during training, while speech can be handled at inference.}
    \label{fig:train_infer_phase}
\end{figure}
\section{Proposed Method}
\subsection{Model Architecture}

TESU-LLM consists of three main components: a Unified Text-Speech Encoder, an Encoder Projector, and a Large Language Model (LLM).
The \textit{Unified Text-Speech Encoder} is designed to produce modality-invariant representations such that semantically equivalent text and speech inputs are mapped to the same latent vector.
This encoder is composed of a \textit{Text Encoder}, a \textit{Speech Encoder}, and a \textit{Unified Text-Speech Mapper}. Formally, the encoders extract intermediate representations: $E_{\text{text}}(x_{\text{text}})$ and $E_{\text{speech}}(x_{\text{speech}})$
These are then aligned into a shared latent space by the mapper \( M(\cdot) \), enforcing the following constraint:
\begin{align}
    M(E_{\text{text}}(x_{\text{text}})) \approx M(E_{\text{speech}}(x_{\text{speech}}))
    \label{eq:mapper}
\end{align}

Our central idea is that by aligning the Unified Text-Speech Encoder with the LLM, we can train the entire system using only text data.
Thanks to the modality alignment enforced by Equation~\ref{eq:mapper}, the model can process speech inputs at inference time as if they were text, without any degradation in performance.
For the Unified Text-Speech Encoder, we adopt SeamlessM4T~\cite{barrault2023seamlessm4t}, an open-source multilingual and multimodal foundation model capable of handling speech-to-text, text-to-speech, and speech-to-speech translation within a single unified architecture.
In our framework, we employ SeamlessM4T’s pretrained text and speech encoders—based on NLLB~\cite{costa2022no} and w2v-BERT 2.0~\cite{chung2021w2v}, respectively.
The unified text-speech mapper is also derived directly from the original decoder module of SeamlessM4T~\cite{barrault2023seamlessm4t}, serving to align modality-specific representations into a shared semantic space.

The \textit{Encoder Projector} is a two-layer MLP that maps the unified latent representation into a space compatible with the LLM’s input embeddings. This output is then concatenated with the LLM’s native token embeddings to form the input sequence for response generation.

During training, only text inputs are used, and representations are computed via \( M(E_{\text{text}}(x_{\text{text}})) \).
At inference time, when TESU-LLM serves as a voice assistant, speech inputs \( x_{\text{speech}} \) are processed through the speech encoder, and the corresponding representation \( M(E_{\text{speech}}(x_{\text{speech}})) \) is used to generate outputs.

\subsection{Training Strategy}
\begin{figure}[t]
    \centering
    \includegraphics[width=\linewidth]{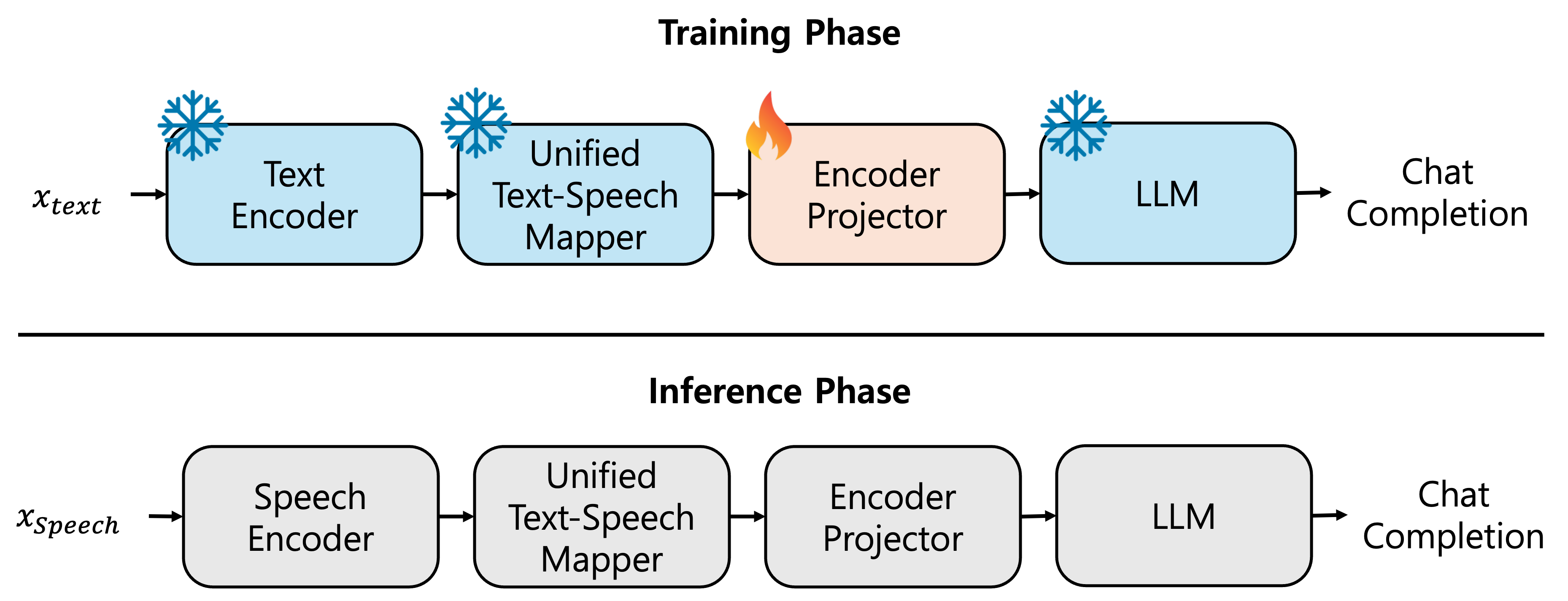}
    \caption{Training (top) and inference (bottom) procedures of TESU-LLM. During training, only text inputs are used. The Text Encoder and Unified Text-Speech Mapper are frozen, and only the Encoder Projector is updated. At inference time, speech inputs are processed through the Speech Encoder and passed through the same mapping and projection modules to generate responses using the frozen LLM.}
    \label{fig:train_infer_phase}
\end{figure}
Fig \ref{fig:train_infer_phase} illustrates the procedures of TESU-LLM for training and inference phase.
We train only the Encoder Projector, while keeping both the Unified Speech-Text Encoder and the pre-trained LLM frozen.
This preserves the alignment between speech and text inputs and maintains the linguistic capabilities of the LLM \cite{wang2024freeze}.
As a result, our method enables efficient training with limited computational resources (e.g., 2 GPUs), while achieving performance comparable to speech-language models trained with significantly more data and compute.

\begin{table*}[t]
\centering
\caption{Comparison of TESU-LLM with recent speech-language models on VoiceBech\cite{chen2024voicebench}. TESU-LLM achieves competitive or superior performance across QA, reasoning, and instruction-following tasks, despite being trained with text-only supervision.}
\label{tab:main_results}
\begin{tabular}{l|cccccccc}
\toprule
\textbf{Model} & \textbf{AlpacaEval} & \textbf{CommonEval} & \textbf{SD-QA} & \textbf{MMSU} & \textbf{OpenBookQA} & \textbf{BBH} & \textbf{IFEval} & \textbf{AdvBench}\\
\midrule
Qwen2-Audio\cite{chu2024qwen2} & 3.74 & 3.43 & 35.71 & 35.72 & 49.45 & 54.7 & 26.33 & 96.73 \\
Moshi\cite{defossez2024moshi} & 2.01 & 1.60 & 15.64 & 24.04 & 25.93 & 47.4  & 10.12 & 44.23 \\
Baichuan-Omni-1.5 \cite{li2025baichuan} & \textbf{4.50}  & \textbf{4.05}  &  43.40  &  \textbf{57.25} & \textbf{74.51} & \textbf{62.7} & \textbf{54.54} & \underline{97.31}\\
Phi-4-multimodal\cite{abouelenin2025phi}  & 3.81 & 3.82 & 39.78 & \underline{42.19} & \underline{65.93} & \underline{61.8}  & 45.35 & \textbf{100.0}\\
GLM-4-Voice \cite{zeng2024glm} & 3.97 & 3.42 & 36.98 & 39.75 & 53.41 & 52.80 & 25.92 & 88.08 \\
LLaMA-Omni\cite{fang2024llama} & 3.70 & 3.46 & 39.69 & 25.93 & 27.47 & 49.2  & 14.87 & 11.35 \\
Freeze-Omni\cite{wang2024freeze} & 4.03 & 3.46 & \underline{53.45} & 28.14 & 30.98 & 50.7  & 23.40 & 97.30 \\
Speechless\cite{dao2025speechlessspeechinstructiontraining} & 3.86 & 2.51 & 35.00 & - & 26.15 & - & - & 62.88\\
\rowcolor{gray!10}
TESU-LLM (Ours)     & \underline{4.17} & \underline{3.91} & \textbf{58.23} & 38.22 & 59.34 & 54.8 & \underline{49.68} & 97.11\\
\bottomrule
\end{tabular}
\end{table*}

\subsubsection{Pretraining Stage}
Inspired by prior works~\cite{wang2025inserter, kim2024paralinguistics, nguyen2025spirit, wang2023blsp}, we adopt an interleaved training strategy that mixes multi-modal encoder outputs with native LLM text embeddings to preserve instruction-following capabilities and mitigate task-specific overfitting.
While interleaved input construction is widely used, many existing approaches rely on extracting precise time-aligned segments from actual speech signals~\cite{nguyen2025spirit, kim2024paralinguistics, wang2023blsp}.
This process is labor-intensive and prone to errors due to co-articulation effects, where phonetic content from neighboring words may bleed into or be truncated from the target segment.
Inserter~\cite{wang2025inserter} addresses this by synthesizing short TTS audio spans for selected word segments, yet this introduces its own challenges such as reliance on TTS quality and increased computational cost.

In contrast, our method eliminates the need for time alignment or speech synthesis altogether.
By leveraging a unified speech-text encoder, we generate speech-like latent representations directly from selected text spans, reducing both alignment artifacts and computational burden.
During pretraining, we randomly sample multiple spans of 3 to 10 consecutive words from long-form text and replace their embeddings with the output of \( M(E_{\text{text}}(x_{\text{text}})) \).
To minimize stylistic variation, the long-form texts are fully regenerated by the base LLM prior to masking, ensuring consistency in writing style~\cite{wang2023blsp, zhang2024dissecting}.
The model is trained with cross-entropy loss computed only on the unmasked text token positions.
This selective supervision encourages the model to integrate speech-conditioned embeddings into context, without disrupting its core language modeling ability.

\subsubsection{Supervised Fine-Tuning}
To enhance instruction-following capabilities as a voice assistant, we perform Supervised Fine-Tuning (SFT) using instruction-style question answering datasets in text format.
During this phase, user queries are encoded using \( M(E_{\text{text}}(x_{\text{text}})) \), while all responses are regenerated by the base LLM to maintain stylistic consistency and alignment with the model's generative capabilities.
In addition to standard instruction data, we introduce a \textit{Repetition Task} similar to \cite{grattafiori2024llama}, guided by a system prompt such as:
\textit{"Reproduce the user's exact query or statement without any interpretation or modification."} This task is designed to enforce precise reproduction of the input embeddings from \( M(E_{\text{text}}(x_{\text{text}})) \), encouraging the model to retain and recover surface-level information accurately.
The same prompt is also employed during evaluation on ASR-style tasks to assess the model's ability to faithfully reconstruct the spoken or encoded input, ensuring consistency between fine-tuning objectives and inference behavior.

\section{Experiments}
\subsection{Datasets}
As described earlier, our entire training pipeline relies solely on text-based data, without the use of any raw speech during training.
For the pretraining stage, we construct a large-scale corpus by sampling the first sentence of each document from OpenWebText\cite{Gokaslan2019OpenWeb} and extending it using generation from a base LLM.
This procedure ensures stylistic consistency and diversity in long-form content, resulting in a total of 200K synthetic samples for pretraining.
For the SFT stage, we collect open-source instruction datasets including UltraChat\cite{ding2023enhancing}, OpenHermes 2.5\cite{OpenHermes2.5}, OpenbookQA\cite{OpenBookQA2018}, and CommonSenseQA\cite{talmor-etal-2019-commonsenseqa}.
These resources provide a wide variety of instruction-response pairs suitable for aligning the model's behavior as a voice assistant.
Additionally, to support the Repetition Task, we utilize the LibriSpeech corpus\cite{panayotov2015librispeech}.
Note that only the textual transcriptions from LibriSpeech are used—no audio data is involved—further reinforcing the text-only training paradigm.
These transcriptions serve as the source of user queries that the model is trained to reproduce verbatim under the repetition objective.
\subsection{Model Configuation}
The base LLM used in our framework is LLaMA-3.1-Instruct 8B,\footnote{\url{https://huggingface.co/meta-llama/Llama-3.1-8B-Instruct}} a publicly available instruction-tuned language model.
For the Unified Text-Speech Encoder, we adopt the open-source SeamlessM4T-v2-large checkpoint,\footnote{\url{https://huggingface.co/facebook/seamless-m4t-v2-large}} which provides robust multilingual and multimodal representation capabilities.
The Encoder Projector is implemented as a two-layer MLP, consisting of only 13 million trainable parameters. Notably, all other components—including the LLM and the Unified Text-Speech Encoder—are kept frozen during training. As a result, the entire training process updates just 13M parameters, enabling highly efficient and resource-conscious optimization.

For both the pretraining and SFT stages, we trained the model for 3 epochs using a batch size of 64 and a learning rate of $1 \times 10^{-4}$.
A cosine learning rate scheduler was applied, with a 3\% warm-up phase.
All experiments were conducted using 2 NVIDIA A100 GPUs with 40GB memory each.
\subsection{Evaluations}
%
To evaluate the model's performance as a voice assistant, we adopted the VoiceBench\cite{chen2024voicebench} benchmark.
VoiceBench is a comprehensive evaluation suite designed to assess the real-world capabilities of LLM-based voice assistants.
It comprises a diverse set of sub-benchmarks, including AlpacaEval, CommonEval, SD-QA, MMSU, OpenBookQA, IFEval, and AdvBench.
These benchmarks collectively measure a range of abilities such as question answering, instruction following, scientific reasoning, and safety alignment, providing a holistic assessment of voice assistant competence.
To evaluate the speech recognition performance of the model, we used the test-clean and test-other subsets of the LibriSpeech dataset\cite{panayotov2015librispeech} and reported results in terms of Word Error Rate (WER).
Additionally, we assessed the model’s ability on the speech-to-text translation (S2TT) task using the CoVoST2 dataset\cite{wang2020covost}, a multilingual corpus that provides aligned speech and text translation pairs across several language directions. For this evaluation, BLEU score was used as the performance metric.
It is worth noting that the S2TT task was not included during any stage of model training, making this evaluation a true zero-shot setting.
For inference, we prompted the model with the instruction: \textit{``Translate the user's query or statement in the \{\texttt{tgt\_language}\}''}, guiding the model to perform multilingual translation in-context based solely on the speech input.

\begin{table}[t]
\centering
\caption{ASR and S2TT performance comparison on standard benchmarks. WER (\textdownarrow) is evaluated on the LibriSpeech test-clean and test-other sets, while BLEU (\textuparrow) is evaluated on CoVoST2 en-de.}
\label{tab:asr_mt}
\begin{tabular}{l|cc|c}
\toprule
\multirow{2}{*}{\textbf{Model}} & \multicolumn{2}{c|}{\textbf{ASR-WER ↓}} & \textbf{S2TT-BLEU ↑} \\
                                & test-clean & test-other & CoVoST2 \\
\midrule
SeamlessM4T-v2\cite{barrault2023seamlessm4t}  & 2.60  & 4.86  & \textbf{37.16} \\
Qwen2-Audio\cite{chu2024qwen2}  & 1.74  & 4.03  & 29.72 \\
Phi-4-multimodal\cite{abouelenin2025phi} & \textbf{1.67}  &  \textbf{3.82} & 34.22 \\
BLSP\cite{wang2023blsp} & 6.40 & - & 23.30 \\
Speechless\cite{dao2025speechlessspeechinstructiontraining}  & 2.47  & 4.65  & - \\
TESU-LLM (Ours)       & 3.61  & 6.20  & 24.30 \\
\bottomrule
\end{tabular}
\end{table}

\section{Results}
\subsection{Speech LLM Benchmark}

We evaluate TESU-LLM on VoiceBench\cite{chen2024voicebench}, a standardized and comprehensive benchmark suite for assessing the instruction-following, reasoning, and factual understanding abilities of speech-enabled LLMs. VoiceBench\cite{chen2024voicebench} aggregates multiple evaluation sets—including AlpacaEval, CommonEval, SD-QA, MMSU, OpenBookQA, BBH, IFEval, and AdvBench—each targeting different aspects of voice assistant capabilities, from general knowledge to robustness against adversarial prompts.
As summarized in Table~\ref{tab:main_results}, TESU-LLM achieves strong overall performance across all categories.
Notably, it obtains the best accuracy on SD-QA (58.23) and ranks second on AlpacaEval (4.17), CommonEval (3.91), and IFEval (49.68).
These results are particularly significant considering that TESU-LLM is trained solely with text-based instruction data, without any paired speech-text supervision.

In comparison, high-performing baselines such as Baichuan-Omni~\cite{li2025baichuan} (e.g., 81.1 on OpenBookQA and 62.7 on BBH), Phi-4-Multimodal~\cite{abouelenin2025phi}, and Qwen2-Audio~\cite{chu2024qwen2} rely on large-scale speech-text datasets and extensive computational resources for training.
Speechless~\cite{dao2025speechlessspeechinstructiontraining} represents the closest approach to ours, utilizing text-only data and avoiding the use of TTS or speech alignment tools.
Despite this similarity, TESU-LLM surpasses Speechless~\cite{dao2025speechlessspeechinstructiontraining} by significant margins in key metrics—for example, +1.16 on AlpacaEval, +1.4 on CommonEval, +22.59 on SD-QA, and +4.33 on IFEval.
This highlights the effectiveness of our unified speech-text encoder alignment framework in capturing semantic information relevant to speech understanding, while maintaining computational efficiency and training simplicity.

\subsection{ASR and Speech Translation}

We evaluate TESU-LLM on two standard speech understanding tasks: ASR and S2TT.
Table~\ref{tab:asr_mt} presents the results in terms of WER for ASR and BLEU score for S2TT.
In the ASR task, TESU-LLM achieves a WER of 3.61\% on \texttt{test-clean} and 6.20\% on \texttt{test-other}, demonstrating that the unified encoder produces sufficiently rich latent representations to support accurate transcription.
While top-performing models like Phi-4-Multimodal\cite{abouelenin2025phi} and Qwen2-Audio\cite{chu2024qwen2} outperform TESU-LLM with WERs below 4\%, these models rely on extensive speech-text supervision and domain-specific training.

For the S2TT task, which was never seen during training, TESU-LLM achieves a BLEU score of 24.30, demonstrating strong zero-shot generalization capabilities.
Unlike models such as Qwen2-Audio\cite{chu2024qwen2} and Phi-4-Multimodal\cite{abouelenin2025phi}, which explicitly fine-tune on S2TT tasks during supervised instruction tuning, TESU-LLM does not require any paired translation data or task-specific adaptation.
While its BLEU score falls short of SeamlessM4T-v2\cite{barrault2023seamlessm4t} (37.16), which is specialized for speech translation, TESU-LLM still outperforms models like BLSP\cite{wang2023blsp} that also exclude S2TT from their training, showcasing the effectiveness of our modality-aligned framework in adapting to unseen speech-language tasks without retraining.

\section{Limitations and Future Work}

While TESU-LLM performs well on core speech understanding tasks, it does not address paralinguistic features such as emotion or prosody, which are essential for more human-aligned spoken LLM.
Future work will extend the framework to incorporate these signals.

\section{Conclusion}

This paper presented TESU-LLM, a novel framework for training speech-capable large language models using only text-based supervision.
By leveraging a Unified Text-Speech Encoder and aligning its output to a pretrained LLM via a lightweight encoder projector, the model can handle spoken queries without requiring any speech data during training.
This approach removes dependencies on paired speech-text datasets, TTS generation, and time-aligned segmentation—challenges commonly faced in previous work.
Extensive experiments on VoiceBench, ASR, and S2TT tasks show that TESU-LLM performs competitively, matching or exceeding models trained with large-scale multimodal data and significant computational resources.
These results highlight the effectiveness of modality alignment in enabling scalable and resource-efficient development of speech-capable language models.

\pagebreak
\bibliographystyle{IEEEtran}
\bibliography{IEEEfull}

\end{document}